# A Bottom Up Procedure for Text Line Segmentation of Latin Script

Himanshu Jain
Department of Computer Science and Engineering
Manipal Institute of Technology
Manipal, India

Archana Praveen Kumar
Department of Computer Science and Engineering
Manipal Institute of Technology
Manipal, India

*Abstract*—In this paper we present a bottom up procedure for segmentation of text lines written or printed in the Latin script. The proposed method uses a combination of image morphology, feature extraction and Gaussian mixture model to perform this task. The experimental results show the validity of the procedure.

I. INTRODUCTION

Extraction of individual separated lines from text images is a vital preprocessing step to many OCR applications like document recognition etc. Though it may seem simple at first glance, the imperfect and varying nature of human handwriting makes it a very challenging task. A few of the problems encountered during segmentation procedures is the difference in skew angle of the different lines, presence of noisy elements like signatures, touching of adjacent lines, overlapping of words etc. Moreover, punctuations marks may lie in between two lines adjacent line making it difficult to classify it into the correct line.

The main highlights of the algorithm include i) its ability to handle unconstrained text ii) its applicability to both handwritten and printed images.

This paper is organized as follows, i) discussion of related work in section 2, ii) description of the preprocessing steps using image morphology in section 3, iii) description of the main segmentation procedure, iv) description of the post-processing steps, and v) Experimental results followed by vi) conclusion.

II. RELATED WORK

A number of line segmentation methods have been proposed to date. The majority of these procedures make use of one of three techniques, i) Hough transforms [1], which make use of different points like gravity centers, local minimas etc. of the connected components to detect possible lines ii) projection profiles [4] which look at the number of elemental pixels across vertically divided strips to identify local maxima and local minima which indicate the presence of a text line and a space between two lines respectively, and iii) smearing methods [6], where a fuzzy run-length is used to segment lines. A measure is calculated for every pixel on the initial image that describes how far one can see when standing at a pixel along horizontal direction. By applying this measure, a new grayscale image is created which is binarized and the lines of text are extracted from the new image.

The above methods however suffer from an obvious drawback, that is their inability to detected lines that are not in a straight line. Therefore, a bottom-up approach is devised which attempts to takes care of the same. Not many bottom-up approaches have been proposed in the literature. One such technique [4] makes using of Minimal spanning tree clustering with distance metric learning. In this technique, a distance metric is introduced to group connected components into tree structure. A set of objective functions are then used to dynamically cut the edges of this tree to extract the text lines. Supervised learning is used in this approach to avoid artificial parameters.

III. PREPROCESSING

A set of preprocessing steps is required to make the data suitable for computation. These steps have been described below:

*A. Binarization*

Given a RGB valued or Gray-scale image, this image is converted into a binary image using different thresholding techniques already proposed in the literature. The binarization technique used in the proposed algorithm is Otsu's thresholding. Otsu's thresholding has been found to produce good results on text data. The image may need cleaning depending on the application domain to remove any noisy elements.

*B. Skew Correction*

Once the binary image has been obtained, a skew correction is performed to make the set of lines as horizontal as possible. To do this, the vertical projection profile is obtained for the image rotated across different angles in the range ($-\pi/4$, $\pi/4$). The standard deviation of these projection profiles is calculated. The one with the maximum standard deviation indicates the optimal rotation angle.

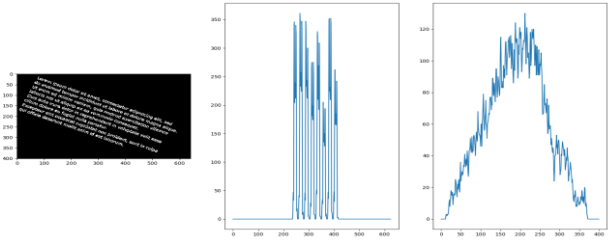

*Figure 1: Projection profiles. (Left: original image, Center: profile at -24 degrees, Right: profile at 0 degrees)*

## C. Extraction of Connected Componets (CC) and Hole filling

Given the rotated binary image, 8-connected components are extracted from the image. These 8-connected components are the smallest units of data used for classification into lines. Once the CCs have been extracted, the holes in each CC are filled as shown in fig. 1. This is done to take care of the variations in handwriting.

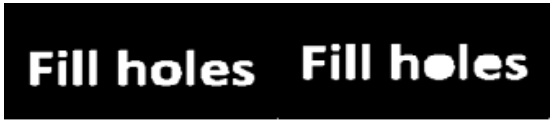

*Figure 2: Hole filling*

## D. Skeletonization/Thinning

Skeletonization is the thinning of connected components to retain the middle-most 1-pixel wide while maintain their basic structure (Fig.3b). A skeletonization procedure is performed on the filled connected components, this is required for the next stage of feature extraction. A variety of procedures have been proposed in the literature for skeletonization. The procedure used should be one that produces the least number of extra branches in the output. The skeleton may also be smoothed to improve the output from the feature extraction stage.

## E. Feature Extraction

Using the skeleton, a set of features are extracted for each CC. The set of features obtained will be used later to classify punctuation marks, noisy components, (components belonging to 2 or more adjacent lines), and also to improve the accuracy of the line of minimum deviation used in the main procedure. The set of features extracted are namely i) strokes, and ii) significant points, which consists of minima points, maxima points, junction points and pen-lift points.

The strokes obtained are the different arcs of the skeleton (figure 3). The stroke extraction starts from the left-most unprocessed pixel of the component and is broken down at every i) junction, ii) end-point and iii) change in vertical direction of stroke arc from increasing to decreasing or vice-versa, iv) change in horizontal direction of stroke arc from increasing to decreasing and vice versa (In fig. 3 notice change in color from (29,26) to (30,25)). This extraction procedure is repeated until all the pixels (except junctions) in the component have been used.

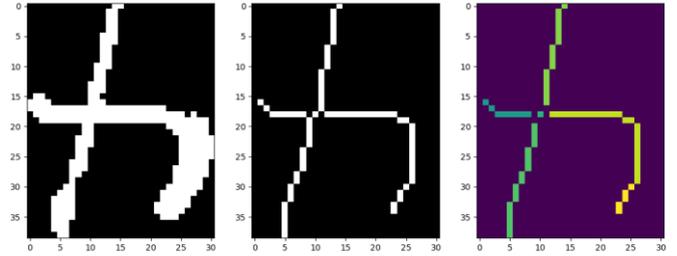

*Figure 3: Stroke Extraction (Left: binary filled image, Center: skeleton image, Right: Strokes labels with 6 strokes)*

## IV. MAIN PROCEDURE/ CC CLUSTERING

After preprocessing is complete, the filled CCs may be grouped or classified into distinct lines/ clusters in a bottom up fashion. The clustering is done starting from the (w.r.t spatial position) left-most CC and ends with the rightmost CC. The measure used for ordering is the horizontal coordinate of the left-most point of CCs.

The clustering is done using a nearest neighbor classification based on the distance measure defined in section IV (A). While clustering, once the rightful cluster has been identified, 3 conditions must be checked for before the CC can be assigned to the cluster. These conditions are defined in Sections IV (C), IV (D), IV (E). IV(C) defines the procedure for check if a component is a special component like a punctuation mark etc. IV(D) defines a set of rules to check if the CC belongs to an uninitialized line/cluster. This step is important because we do not specify the total number of lines/ clusters beforehand but initialize them dynamically during the procedure. Finally, if the CC is neither a special component nor a part of an uninitialized cluster, the procedure described in IV (E) is performed to check the CC contains two touching words belonging to different lines. Section IV (B) describes a Gaussian mixture of heights that will be used during the whole procedure.

## A. Distance Measure

Before the distance of a CC from the different clusters is calculated. A set of 5 candidate clusters must be identified using the Euclidean distance between the closest pair of points in the cluster and the CC. Brute force technique provides incredibly poor performance for this problem. Therefore, the Euclidean distance between the point of the line horizontally closest to the left-most point of CC and the left-most point of the CC itself is taken as a measure to find the candidate clusters. The clusters that produce the minimum distance measure as chosen as our candidate clusters. This pair may not be the closest pair between the cluster and CC but is a good substitute for our procedure. Once the candidate clusters have been identified, the closest cluster is identified using the distance measure. The distance measure is defined as the combined deviation of the points of the CC from a set of regression lines. A regression line is a line of the form (1) which is defined for a cluster every time a CC is added to it. This regression line is defined using (2).

$Y = slope * X + intercept$ (1)

Where slope, intercept are constants.

slope=0, intercept=(top+bottom)/2 if count(sign_points)<25

slope=m, intercept=c if count(sign_points)>=25 (2)

where top, bottom are the highest and the lowest vertical co-ordinates of the CC respectively. Count(x) is the number of points in the set x, sign_points is the set of significant points of the CCs belonging to the cluster. And m, c are the coefficients of the line of minimum deviation from 25 of the rightmost significant points of the CCs in the cluster.

The number of regression lines corresponding to a cluster at any point of time must be limited to 5. The selected regression lines are the ones with the minimum deviation from the points of the CC added last to the cluster. This limiting of the number of regression lines makes sure that only the right-most neighborhood of a cluster is used during computation. Thus, accounting for variations in the skew along the line.

*B. Gaussian Mixture of Heights*

For each cluster, the height of the ascenders and descenders like h, g, l, p etc. must be distinguished from the height of normal letters like a, c, e etc. To do this a window of width 1 is examined over the width of each CC in the cluster as show in fig. 4. The set of highest vertical coordinate of the CC minus the lowest vertical coordinate for each window is fitted to a Gaussian Mixture Model having 3 components/classes using the EM algorithm. Physically, these 3 components represent the heights of i) ascenders and descenders like f, g, ii) normal components like a, c, and iii) lower heights corresponding to windows having only one block of white pixels. The means of the 3 Gaussian components will be further references as CLUSTER_HT1, CLSUTER_HT2, and CLUSTER_HT3 ordered from highest to lowest.

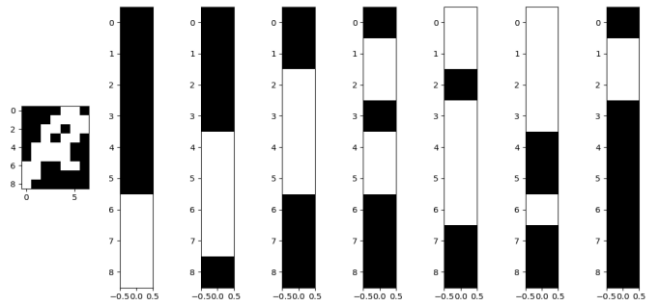
Figure 4: Windows for Height classification

*C. Check for Special Components*

A CC is added to a list of special components if any one of a set of conditions is satisfied. The CCs belonging to this list are classified to clusters in a post-processing step. This list consists of any punctuation marks or other components that cause an undesirable change in the slope of the regression lines. These are the components that lie above and below the actual lines of text, for example, the dot in an 'i', quotation marks, accent marks in Greek script etc. This step is therefore language specific. The proposed procedure was designed for the English language originally and therefore, a set of 2 rules were formulated as in (3) which takes care of two types of special components, i) small and simple components like dots and commas (specified in the first part of the OR statement in (3)), and ii) a disconnected horizontal line on top of 'T', 'I' (specified in the second part of the OR statement (3)).

(Count(strokes)<5 AND HT<CLUSTER_HT2) OR
(count(strokes)<5 AND HT<3*WD)  (3).

Where, count(x) is the number of item in the set x, strokes is the set of extracted strokes of the CC, HT is the height of the CC, CLUSTER_HT2 is the 2nd largest mean of the classes defined in Section IV (B)

*D. Check for New Cluster*

After the closest cluster has been identified and it has been established that the component is not a special component, the CC may be added either to a new cluster, i.e. a new line or to the closest cluster. 2 rules are defined to perform the right action. The CC is assigned to a new cluster if either of the rules is satisfied. i) the ratio of the area of the CC in the window of width CLUSTER_HT1 across the regression line of the closest cluster having the least deviation from the CC to the total area of the CC is less than 0.2 or ii) the number of strokes of the CCs in the closest cluster and the CC overlapping in the horizontal direction is greater than 3 for both the cluster and the CC. This second rule is derived specifically for Latin based languages, based on the idea that no two words in a line overlap in the vertical direction.

Once it has been established that a CC must be a part of a new cluster, a component break procedure as described in 4E must be performed using the closest cluster as the test cluster and the CC as the test CC. This is to check for overlapping of words across the closest cluster/line and the new cluster/line. After a CC has been classified to a new cluster, it must check for already processed CCs to be a part of the new cluster. This is needed for those cases where the first word of a line touches a word of a different line. To do so, a measure MAX_GAP is defined as in (4). We then, examine the already classified CCs sorted from right to left using their centers of gravity. If the minimum Euclidean distance between the new cluster and the classified CC being examined is less than MAX_GAP and the height of the CC is greater than CLUSTER_HT1 of the closest cluster, then, the component break procedure (described in 4E) is performed with the new cluster as the test cluster and the CCs being examined as the test CC.

$MAX\_GAP = 3 * CLUSTER\_HT1$ (4)

Where CLUSTER_HT1 is the largest mean of the classes in the Gaussian mixture of heights in the cluster. It is to be noted that this parameter is not optimal and requires further improvement.

### E. Component Break Procedure

The component break procedure takes one test cluster and one CC as input and tests if the test CC is a part of the test cluster. To do so, a mean regression line is calculated for the test cluster. This mean regression line is calculated as i) the mean of right regression lines in the cluster which were described in (2) if the CC is being examined to be a part of an already initiated cluster, and ii) the mean of left regression lines as described in (5) if the CC is being examined to be a part of a newly initiated cluster, i.e. if the procedure is called from 4C.

slope=0, intercept=(top+bottom)/2 if count(sign_points)<25

slope=m, intercept=c if count(sign_points)>=25          (5)

where top, bottom are the highest and the lowest vertical co-ordinates of the CC respectively. Count(x) is the number of points in the set x, sign_points is the set of significant points of the CCs belonging to the cluster. And m, c are the coefficients of the line of minimum deviation from 25 of the left-most significant points of the CCs in the cluster.

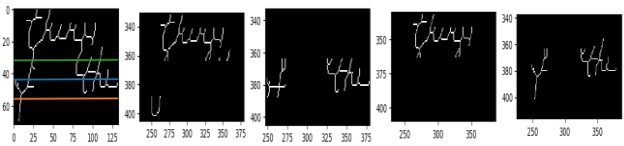

*Figure 5: Breaking of components (From left to right: a) skeleton of original component with blue mean regression line, b) skeletons of out-components before correction c) skeletons of in-components before correction, d)skeleton of out-components after correction, v) skeleton of in-components after correction*

A window of width equal to CLUSTER_HT2 of the test cluster is examined across the regression line as shown in fig. 5a. The test CC is broken into a set of in-components, the CC that occur inside a window and a set of out-components, the CC that occur outside. If any CC in the set of in-components has more than 4 strokes, the CC must be broken. Else, there is no need for break. Before breaking, a correction must be performed to take care of faulty breaks in ascenders and descenders (fig. 5c). This is done by moving the components in in-components having number of strokes less than 4 to the set of out-components and vice-versa.

## V. POST-PROCESSING

After the lines have separated, a series of post-processing steps must be followed to produce the final output. These steps are described in the subsections that follow:

### A. Combine Clusters

The main procedure of component clustering may create a few false clusters due to the use of artificial parameters. These false clusters must thus be eliminated. These false clusters are grouped using the following procedure.

First, for each cluster a median line is defined, that connects the centers of gravity of all the adjacent CCs in that cluster. This median line is extended to the left-most and right-most horizontal co-ordinates of the cluster. While combining two lines, two cases follow: i) the two clusters overlap vertically, ii) one cluster exists completely to the right of the other. In case (i), the overlap region is examined and the absolute height difference HT_DIFF between the median lines of the 2 clusters in this region is calculated. The vertical overlap between the two clusters is calculated as OVERLAP. Two clusters i and j following case (i) are then combined if both (6) and (7) follow.

(HT_DIFF<CLUSTER_HT1i)                     AND
(HT_DIFF<CLUSTER_HT1j)                      (6)

OVERLAP<0.33*CLUSTER_WDi                    OR
OVERLAP<0.33*CLUSTER_WDj                    (7)

Where CLUSTER_HT1i, CLUSTER_HT1j are the max. means of classes in the Gaussian Mixture of heights in cluster i and cluster j respectively. CLUSTER_WDi, CLUSTER_WDj are the width of cluster i and cluster j respectively.

If case (ii) follows, HT_DIFF is calculated as the difference between right-most vertical coordinate of the median line of the left cluster and left-most vertical coordinate of the median line of the right cluster, and the two clusters i, j are combined if (5) follows. Refer Fig. 6.

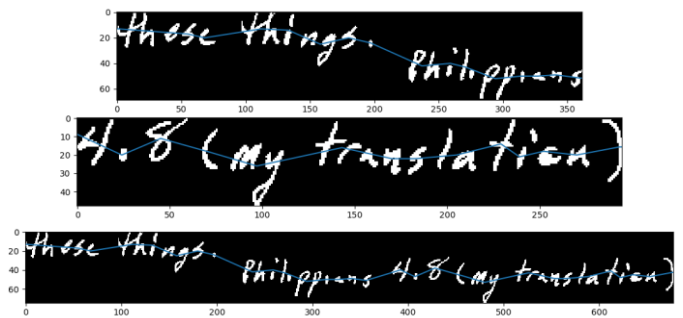

*Figure 6: Combining Clusters/Lines*

### B. Classification of Special Components

An item in the list of special components is classified to the cluster having the minimum Euclidean distance between the closest pair of points in the cluster and the special component.

At the end of the post-processing stage, each cluster represents one line.

## VI. EXPERIMENTAL RESULTS

Experiments were performed on 400 images with 7641 lines selected from the ICDAR 2009 text segmentation contest dataset [11], ICFHR 2010 text segmentation contest dataset benchmarking dataset [16], and ICDAR 2013 text segmentation contest dataset [17]. These include document images written in Greek, French, German and English and not all datasets are disjoint. We used a set of measures commonly used in the literature and in the evaluation of the ICDAR contests. These measures are defined in [18]. N is the total number of lines present in the dataset, M is the number of identified lines, DR is the Detection Rate, RA is the Recognition Accuracy, FM is the Final Measure. The experimental results are tabulated below:

*Table 1: Experimental Results (w/o rotation)*

| M_THRESH | N | M | DR | RA | FM | o2o |
|---|---|---|---|---|---|---|
| 95% | 7641 | 7675 | 95.91% | 95.49% | 95.70% | 7329 |
| 90% | 7641 | 7675 | 98.78% | 98.34% | 98.56% | 7548 |

The table clearly proves the validity of the algorithm giving an FM measure of 95.70% with a base threshold of 95%. It must be noted that this procedure was built specifically for the English language, but still produces good results for other languages in the Latin script. It must also be noted that the evaluation method does not allow for the rotation stage during preprocessing. The results obtained are, therefore, sub-optimal. Visual inspection of the line segments after rotation revealed much better results as expected.

Apart from the aforementioned datasets, the algorithm was also tested on a set of 62 images downloaded from Google and a set of 100 images from the IAM database [3] containing both handwritten and printed text images in English. Visual inspection of the output of the procedure on these images produced good results at handling unconstrained text. Illustration of the segmentation on some samples is shown in fig.7.

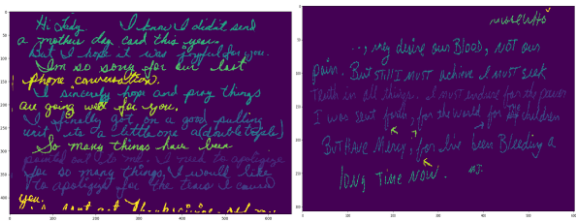

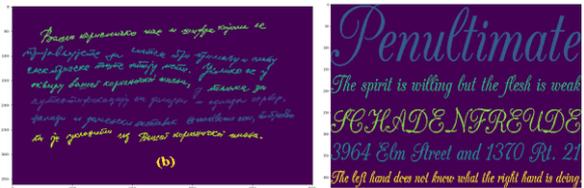

*Figure 7: Illustration of the proposed line segmentation procedure for samples with varying skew angles, sizes etc.*

The main highlight of this algorithm is its ability to handle unconstrained text. While the main drawbacks are its intensive computation and use of multiple artificial parameters.

## VII. FUTURE WORK AND CONCLUSIONS

This paper proposed a line segmentation methodology using a bottom-up approach which was based on image morphology and feature extraction. The algorithm performed significantly well on printed documents and handwritten documents with well-separated lines and moderately well on document containing overlapping words. The main advantage of this algorithm is its ability to detect lines across varying samples. The main disadvantage is its use of multiple artificial parameters.

Future work, thus, includes improvement in the breaking procedure performed on overlapping words, introduction of machine learning techniques to eliminate the use of artificial parameters and incorporation of the optimal rotation angle into the algorithm.